\pdfoutput=1

\documentclass[11pt]{article}

\usepackage[]{acl}

\usepackage{times}
\usepackage{latexsym}

\usepackage[T1]{fontenc}

\usepackage[utf8]{inputenc}

\usepackage{microtype}

\usepackage{inconsolata}

\usepackage{graphicx}
\usepackage{subcaption}
\usepackage{multirow}
\usepackage{tabularx}
\usepackage{todonotes}
\usepackage{amsmath}
\usepackage{booktabs}
\usepackage{tcolorbox}
\usepackage{mathtools}
\usepackage{amsfonts}
\usepackage{bbm}

 \newcommand*\iftodonotes{\if@todonotes@disabled\expandafter\@secondoftwo\else\expandafter\@firstoftwo\fi}  %
\makeatother

\newcommand{\mathtutorbench}{\textsc{MathTutorBench}}

\newcommand{\vparam}{\vtheta}

\newcommand\cut[1]{}

\newcommand{\squishlist}{
   \begin{list}{$\bullet$}
    { \setlength{\itemsep}{0pt}      \setlength{\parsep}{3pt}
      \setlength{\topsep}{3pt}       \setlength{\partopsep}{0pt}
      \setlength{\leftmargin}{1.5em} \setlength{\labelwidth}{1em}
      \setlength{\labelsep}{0.5em} } }

\newcommand{\squishlisttwo}{
   \begin{list}{$\bullet$}
    { \setlength{\itemsep}{0pt}    \setlength{\parsep}{0pt}
      \setlength{\topsep}{0pt}     \setlength{\partopsep}{0pt}
      \setlength{\leftmargin}{2em} \setlength{\labelwidth}{1.5em}
      \setlength{\labelsep}{0.5em} } }

\newcommand{\squishend}{
    \end{list}  }

{}
{}
{}
{}

\newcommand{\real}{\mbox{$\mathbb{R}$}}

\newcommand{\myvec}[1]{\mbox{$\mathbf{#1}$}}
\newcommand{\myvecsym}[1]{\mbox{$\boldsymbol{#1}$}}

\newcommand{\vtheta}{\mbox{$\myvecsym{\theta}$}}

\newcommand{\va}{\mbox{$\myvec{a}$}}

\newcommand{\vi}{\mbox{$\myvec{i}$}}

\newcommand{\vp}{\mbox{$\myvec{p}$}}
\newcommand{\vq}{\mbox{$\myvec{q}$}}

\newcommand{\vs}{\mbox{$\myvec{s}$}}

\newcommand{\vu}{\mbox{$\myvec{u}$}}

\newcommand{\vx}{\mbox{$\myvec{x}$}}

\newcommand{\vy}{\mbox{$\myvec{y}$}}

\newcommand{\be}{\begin{equation}}
\newcommand{\ee}{\end{equation}}
\newcommand{\bea}{\begin{eqnarray}}
\newcommand{\eea}{\end{eqnarray}}
\newcommand{\beaa}{\begin{eqnarray*}}
\newcommand{\eeaa}{\end{eqnarray*}}

\newcommand{\blockcomment}[1]{}

\makeatother

\title{MathTutorBench: A Benchmark for Measuring Open-ended\\ Pedagogical Capabilities of LLM Tutors}

\author{
    Jakub Macina$^{1, 2}$ \quad
    Nico Daheim$^{1, 3}$ \quad
    Ido Hakimi$^{1, 2}$ \quad \\
    \textbf{
    Manu Kapur$^4$ \quad
    Iryna Gurevych$^3$ \quad
    Mrinmaya Sachan$^{1}$
    } \\ \text{} \\
  $^{1}$Department of Computer Science, ETH Zurich \quad
  $^2$ETH AI Center \\
  $^{3}$Ubiquitous Knowledge Processing Lab (UKP Lab), Department of Computer Science, \\ Technical University of Darmstadt and National Research Center \\ for Applied Cybersecurity ATHENE, Germany \\
  $^4$Professorship for Learning Sciences and Higher Education, ETH Zurich
}

\begin{document}
\maketitle
\begin{abstract}
Evaluating the pedagogical capabilities of AI-based tutoring models is critical for making guided progress in the field. Yet, we lack a reliable, easy-to-use, and simple-to-run evaluation that reflects the pedagogical abilities of models.
To fill this gap, we present \mathtutorbench, an open-source benchmark for holistic tutoring model evaluation. \mathtutorbench\ contains a collection of datasets and metrics that broadly cover tutor abilities as defined by learning sciences research in dialog-based teaching. To score the pedagogical quality of open-ended teacher responses, we train a reward model and show it can discriminate expert from novice teacher responses with high accuracy.
We evaluate a wide set of closed- and open-weight models on \mathtutorbench\ and find that subject expertise, indicated by solving ability, does not immediately translate to good teaching.
Rather, pedagogy and subject expertise appear to form a trade-off that is navigated by the degree of tutoring specialization of the model.
Furthermore, tutoring appears to become more challenging in longer dialogs, where simpler questioning strategies begin to fail.
We release the benchmark, code, and leaderboard openly
to enable rapid benchmarking of future models.\footnote{Correspondence to \texttt{jakub.macina@ai.ethz.ch}.}

\end{abstract}

\hspace{0em}\includegraphics[width=0.9em,height=0.9em]{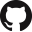}\hspace{.25em}\parbox{\dimexpr\linewidth-2\fboxsep-2\fboxrule}{\href{https://github.com/eth-lre/mathtutorbench}{github.com/eth-lre/mathtutorbench}}

\section{Introduction}
Large Language Models (LLMs) present an opportunity to transform education by offering ubiquitous access to individualized tutoring~\citep{jurenka2024towards}.
\begin{figure}[ht]
    \centering
    \includegraphics[width=0.75\linewidth]{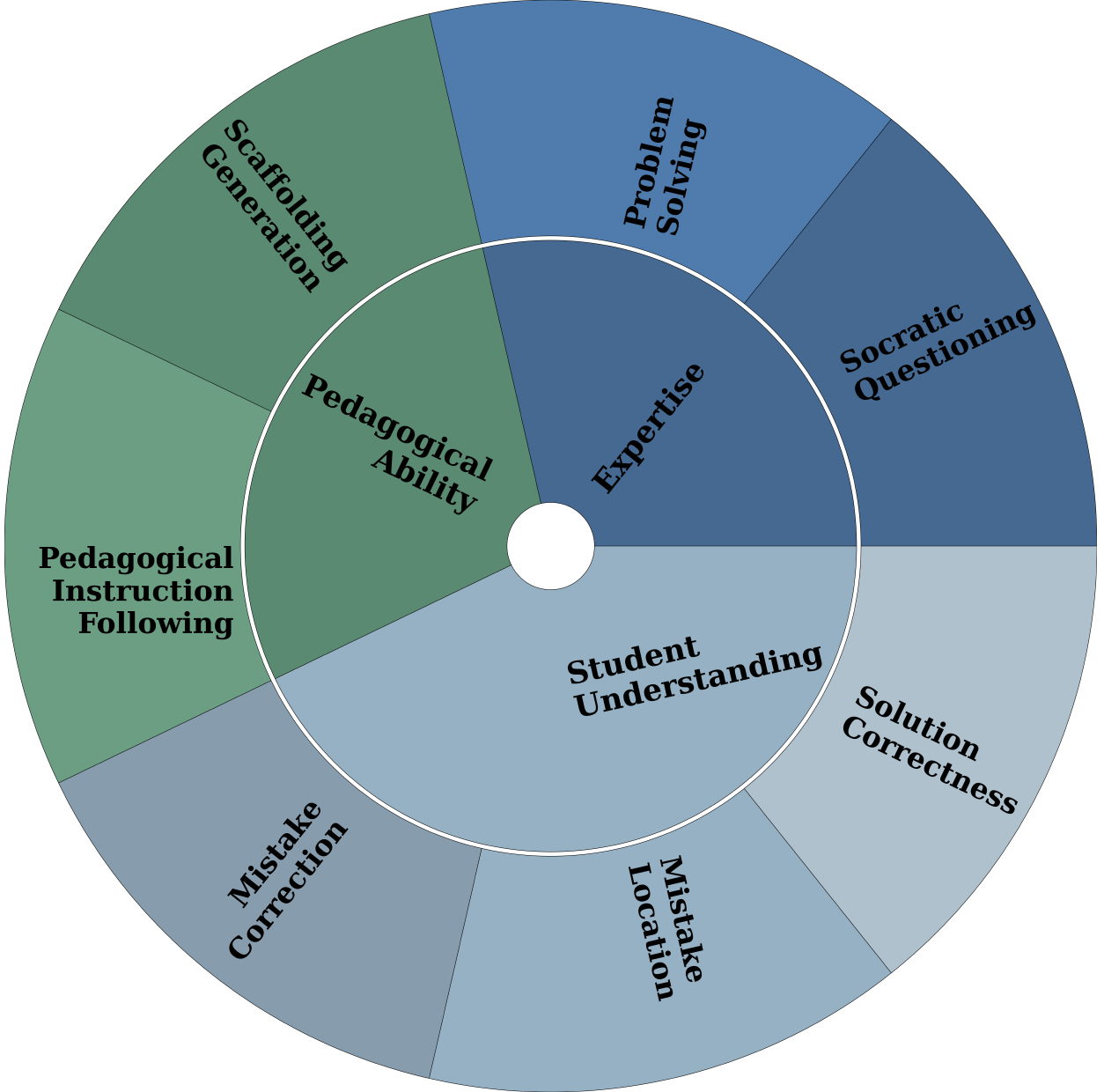}
    \caption{
    Effective teaching requires various skills which we categorize into expertise, student understanding, and pedagogical ability. \mathtutorbench\ evalutes these according to the tasks shown in the outer ring.
    }
    \label{fig:skills-figure}
\end{figure}
While these models excel at generating correct answers~\cite{wei2022chain,achiam2023gpt}, experienced teachers 
help students think for themselves and do not just give away answers to make learning effortless~\cite{sharma2024towards}.
Teaching involves a combination of skills including subject expertise, the ability to diagnose and correct student mistakes, and the application of sound pedagogical techniques. For example, teachers need to know when to withhold answers from students, use Socratic questioning~\cite{anghileri2006scaffolding}, or how to engage them cognitively in problem solving~\cite{chi2014icap,kapur2016examining}.
Therefore, a crucial element of building LLM tutors is their evaluation; it is critical to understand whether their guidance is helpful to prevent harm, and to guide progress in future model development.

Yet, current evaluation practices do not meet these criteria.
On the one hand, automatic metrics usually evaluate tutoring models by measuring the word overlap between a ground-truth response and a generated response~\cite{beasharedtask2023}, or focus exclusively on question-answering performance~\cite{tutorchat24}. This is fast but arguably fails to capture the intricacies of tutoring.
Although human evaluation might be a way to capture these nuances by defining suitable criteria to capture them~\citep{tack2022ai,mrbench2024}, it is expensive. %
Importantly, it can only create a snapshot of current performance and cannot be used to evaluate or compare to future models.

In this work, we fill this gap by releasing \mathtutorbench, a collection of datasets and metrics to holistically evaluate dialog tutoring models for math tutoring. Teaching is a complex and multifaceted task that extends beyond subject mastery~\cite{cognitive_motivation_balance_2008,nye2014autotutor,beasharedtask2023,bridge24}.
Therefore, \mathtutorbench\ is divided into three categories: \emph{math expertise} which evaluates the subject-matter expertise of the tutor, \emph{student understanding} which evaluates the tutor's ability to verify, locate and correct student solutions, and \emph{teacher response generation} which evaluates the scaffolding abilities of the tutor.
Math expertise and student understanding are evaluated based on standard metrics, and we propose a novel metric for evaluating teacher response generation.
In particular, we train a small and quick-to-run reward model by contrasting effective and less effective tutor utterances in terms of structured scaffolding guidance with questions and hints instead of giving away the answer~\citep{anghileri2006scaffolding}. 
The reward model is then used to score tutor model generations. 
We show that this metric is reliable by showing that it can distinguish utterances from expert teachers from those stemming from novice teachers~\citep{bridge24} with high accuracy.

We evaluate various open- and closed-weight state-of-the-art LLMs and specialized tutoring models on \mathtutorbench.
Our results show that there is a trade-off between subject expertise and pedagogical abilities that is dependent on the degree of specialization of a tutoring LLM.
Specializing an LLM for pedagogy comes at the cost of solving ability and, conversely, a high solving accuracy often means that the LLM lacks pedagogy.
Still, more specialized tutoring models tend to retain their teaching abilities even further into a dialog with a student, while general models quickly become worse.
With this, our work contributes to accelerating the development of tutoring LLMs by providing a holistic benchmark that can be evaluated quickly and fairly using automatic metrics.
We release our code and data publicly to promote open research on tutoring LLMs.

\begin{figure*}[t!]%
	\centering
	\resizebox{.99\textwidth}{!}{\includegraphics{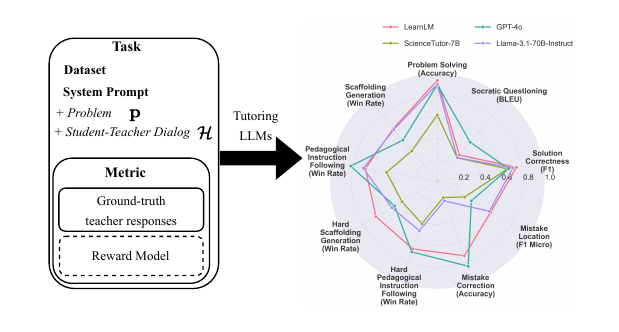}}
	\caption{Overview of the \textit{MathTutorBench} benchmark. Each benchmark task defines a dataset, system prompts with problem and dialog, metric, and ground-truth teacher responses. A reward model is used to score the pedagogical quality over teacher responses (win rate). The right part of the figure shows the outcome as a performance comparison of selected LLMs. While they all perform well in a simple problem-solving setting, most of them lack in correct detection of mistakes and generating pedagogical responses. 
    }\label{fig:overview}
\end{figure*}

\section{Related Work}

\subsection{LLM-Based Dialog Tutoring}
A good tutor should scaffold student learning in a structured way rather than just provide correct answers. 
Current approaches to dialog tutoring using LLMs try to achieve this by different means: prompt-based elicitation of pedagogical behavior, finetuning models on pedagogical conversations, and alignment with pedagogical preferences. 

First, most existing works use LLMs with a carefully chosen prompt which enumerates desired pedagogical behavior. Bridge~\cite{bridge24} analyzes the teacher behavior and proposes to structure the prompt into a sequence of decisions, similar to real teachers, first to infer the error type and then to determine the pedagogical intent. Other works mostly directly write extensive prompts~\cite{class2023,kargupta-etal-2024-instruct}. However, defining such prompts is tedious, sensitive to small changes and difficult to test~\cite{jurenka2024towards}.

Second, several approaches finetune models on real or mostly synthetically generated data.
SocraticLM~\cite{socraticlm2024} uses a GPT-4 judge to evaluate the quality of teacher guidance using correctness and Socratic principles. A similar approach is to role-model teacher and student conversations based on textbook data~\cite{tutorchat24,book2dial2024}. MathDial~\cite{mathdial2023} is one of the few works that use teachers' utterances when interacting with students to finetune models. However, it is expensive to collect such data on a larger scale. Therefore, LearnLM~\cite{team2024learnlm}
uses an empirically validated mixture of synthetic and teacher-created datasets. However, for capturing high-quality tutoring, teacher-created data is essential and therefore upweighted in their final data mix.
Finally, LLMs can be aligned for pedagogical preferences during post-training~\cite{team2024learnlm}, because these are usually tacit. However, no datasets are openly available or they rely on larger models such as GPT-4 as a judge which limits its generalizability.

Our benchmark contributes an important missing ingredient in the development of LLM-based tutors -- the ability to quickly evaluate and compare models on key pedagogical aspects.

\subsection{Automatic \& Human Evaluation}
Several works rely on automatic NLG metrics %
such as BLEU~\cite{papineni2002bleu} or BERTScore~\cite{Zhang2020BERTScore} for evaluation which require human-annotated ground truths. However, since tutoring has the goal of helping students learn %
~\cite{opportunities2023}, it is %
very open-ended and there is no single best pedagogical approach at each turn~\cite{jurenka2024towards}. This results in noisy and unreliable scores from automatic metrics~\cite{beasharedtask2023}. There exists an educational-specific classifier of active teacher listening~\cite{uptake2021}, however, it is limited to only this one dimension of teaching and does not account for the entire dialog history.
Therefore, recent finetuned tutoring LLMs~\cite{tutorchat24,socraticlm2024} rely on GPT-4-as-a-judge on dimensions like helpfulness and presentation. Some works on reasoning~\cite{mathchat2024} also focus on multi-turn model abilities judged by GPT-4 but they lack an educational focus. %

Pedagogical quality annotation requires hiring teachers, but it is time-consuming and hard to compare across trials. Two papers recently addressed the issue~\cite{tack2022ai,mrbench2024} by providing evaluation taxonomies but only present one-off static snapshots of current models' performance without the possibility to automatically evaluate new models. 
Finally, measuring student learning gains directly focuses on the end goal. However, learner studies are costly, time-consuming~\cite{schmucker2024ruffle}, and with strict ethics and privacy requirements. There is a growing interest in designing proper evaluation guidelines~\cite{beasharedtask2023,jurenka2024towards}, however, there is still a need for a unified automatic evaluation for scalable model development. 

\mathtutorbench\ addresses the limitations of existing automatic metrics by focusing specifically on tutoring, is simple to run, replicable, and could serve as a proxy for deciding which models to focus on in human studies. Moreover, our benchmark only uses data collected from real teachers.

\begin{table*}[h]
    \centering
    \resizebox{1.\linewidth}{!}{
        \begin{tabular}{l|cc|ccc|cc}
            & \multicolumn{2}{c}{\textbf{Math Expertise}} & \multicolumn{3}{c}{\textbf{Student Understanding}} & \multicolumn{2}{c}{\textbf{Teacher Response Generation}} \\
            \hline
             \textbf{Task(s)} & Problem  & Socratic  & Solution  & Mistake  & Mistake  & \multicolumn{1}{c}{Scaffolding Gen.,} & \multicolumn{1}{c}{Scaffolding Gen.,} \\
             & solving & questioning &  correctness &  location &  correction & \multicolumn{1}{c}{Ped. instr. following} & \multicolumn{1}{c}{Ped. instr. following}  \\
            \hline
            \textbf{Dataset} & GSM8k & GSM8k & StepVerify &  StepVerify & StepVerify & MathDialBridge & MathDialBridge[hard]  \\
             \textbf{Input} & $\vp$ & $\vp$ & $\vp$, $\widehat{\vs}$ & $\vp$, $\widehat{\vs}$ & $\vp$, $\mathcal{H}$ & $\vp$, $\mathcal{H}$ & $\vp$, $\mathcal{H}$  \\
            \textbf{Type} & generation & generation & bin. clas. (bal.) &  multi-cl. & generation & generation & generation  \\
         \textbf{Ground Truth} & $\va$ & $\vq_1,\dots,\vq_N$ & $\mathbbm{1}(e\neq 0)$ & $e$ & $\va$ & $\vu_{teacher}$ & $\vu_{teacher}$   \\
        \textbf{Instances} & 1319 & 1319 & 2004 &  2004 & 1002 & 1150 & 327   \\
        \textbf{Avg. turns} & - & - & - &  - & 3.04 & 3.08 & 5.78   \\
            \hline
        \end{tabular}
    }
    \caption{Datasets used in the benchmark and their statistics. Notation defined in Section~\ref{sec:background}. }\label{tab:datasets_in_benchmark}
\end{table*}

\section{Background}\label{sec:background}
\subsection{Next Teacher Utterance Generation}

We focus on educational dialogues between a student and a teacher, where a student is trying to solve a multi-step problem $\smash{\vp\in \mathcal{V}^\ast}$. %
The problem has a single numerical solution $\va$ and a sequence of solution steps $\smash{\vs=(\vs_1, \dots, \vs_N)}$, where each $\smash{\vs_n \in\mathcal{V}^\ast}$ and $\smash{\vs_N}$ contains the final answer
$\va$. 
A student solution consists of steps $\smash{\widehat{\vs}=(\widehat{\vs}_1, \dots, \widehat{\vs}_M)}$ and the first step with a mistake is $e \in \{0, 1,\dots, M\}$, where $e=0$ means no mistake. 

The goal of dialog tutoring is to continue an existing teacher-student dialog $\mathcal{H} \coloneqq (\vu_1,\dots,\vu_{T-1})$ consisting of $T-1$ turns $\vu_{t} \in \mathcal{V}^\ast$ with a new turn $\vu_{T} \in \mathcal{V}^\ast$ that simulates the teacher and guides the student towards solving a problem.
This is usually done by learning a model
$p_{\text{\vparam}}(\vu_T\mid \mathcal{H}, \mathcal{K}, \vi_t)$ with parameters $\vparam \in \real^d$, which is optionally conditioned on background knowledge $\mathcal{K}$ (in our case only the problem $\vp$) and a teacher intent $\vi_T$, and using a decoding strategy, such as greedy decoding or sampling according to $p_{\text{\vparam}}$ to generate an output.
The turn $\vu_T$ should then fulfill the desiderate laid out in Section~\ref{sec:ls_principles}.
The goal of this work is to present a benchmark to understand the quality of various $\vu_T$ generated by different models $\vparam$.

\subsection{Learning Sciences Principles}\label{sec:ls_principles}
We focus on 1:1 multi-turn teacher-student interactions where teachers promote active learning~\cite{freeman2014active} by engaging students through scaffolding nudges, hints, and Socratic questioning. Based on effective teaching research~\cite{graesser1995collaborative,LEPPER2002135,litman2006correlations,cognitive_motivation_balance_2008,chi2014icap,nye2014autotutor,jurenka2024towards}, we define the following pedagogical principles: \textit{(a) correctness:} the teacher should guide the student towards the correct answer and not state incorrect facts; \textit{(b) scaffolding instead of giving away the answer:} the teacher should help the student to cognitively engage with the problem and discover the answer on their own; \textit{(c) encourage self-correction:} by correctly identifying the student mistake and first giving the student the opportunity to self-correct and learn from a mistake; \textit{(d) not overload student:} manage cognitive load by not giving too much information at once.

\section{MathTutorBench} %
We introduce \mathtutorbench, a benchmark that evaluates the tutoring capabilities of tutoring models. \mathtutorbench\ consists of three high-level skills that a good human teacher needs to have~\cite{bommasani2021opportunities}: \textit{Expertise}, \textit{Student Understanding} and \textit{Pedagogical Abilities}. These skills are tested by seven different tasks, each consisting of a dataset, prompt, and metric. All tasks in \mathtutorbench\ are related to math tutoring. The problems are mostly sourced from GSM8k~\cite{cobbe2021gsm8k}. 
Table~\ref{tab:datasets_in_benchmark} summarizes the datasets and tasks. The prompts used for the tasks in the benchmark are shown in Appendix~\ref{sec:appendix-task-prompts}.

\subsection{Tasks}
This section explains each task and complements Table~\ref{tab:datasets_in_benchmark} with the rationale for including it.

\noindent\textbf{1. Problem Solving.} We %
include a math word problem solving task that measures the accuracy of the final numeric answer generated with chain-of-thought~\cite{wei2022chain} compared to the answer $\va$. Even though this type of evaluation is popular, saturated, and contaminated, in \mathtutorbench\ it serves as an indicator of a balance between expertise and pedagogical abilities.

\noindent\textbf{2. Socratic Questioning.} Socratic questioning is related to the problem decomposition to smaller and more manageable parts.  
This task is to evaluate whether a model generates for each correct step $\vs_n$ at least one corresponding guidance question $\vq_n$ towards the correct answer, which could be posed to the student instead of simply providing the answer~\cite{shridhar2022automatic,socraticlm2024}. 

\noindent\textbf{3. Student Solution Correctness.}
This task evaluates a teacher's ability to verify the correctness of a student's answer. %
Framed as a balanced binary classification 
task based on student solution chain $\widehat{s}$, this dimension ensures that the model can objectively discern whether a student's reply is correct or incorrect, a crucial prerequisite for providing accurate feedback and identification of misconceptions~\cite{bridge24}.

\noindent\textbf{4. Student Mistake Location.}
Mistake location is a critical component of effective tutoring, focusing on a teacher's ability to accurately identify the exact location of the first mistake in a student's response $\widehat{\vs}$~\cite{verifiers2024}. This task assesses whether a tutoring model can pinpoint where a student's reasoning has gone wrong, enabling timely and precise feedback. By detecting steps with mistakes, the model can help students understand their misconceptions and steer the conversation to mitigate them, thus fostering a more productive learning experience~\cite{kapur2016examining,bridge24}.

\noindent\textbf{5. Student Mistake Correction.}
This task measures the performance of a model to generate a reasoning chain with a correct final numeric answer $\va$ even though the student proposes an incorrect answer in the dialog history $\mathcal{H}$. The conditioning on dialog history is the difference to Problem Solving. We test the models' ability to handle incorrect solutions. Models should not get derailed by students' incorrect steps. From a broader perspective, even if there is an incorrect step in a dialog history $\mathcal{H}$, this tests the recovery of a model from mistakes.    

\noindent\textbf{6. Scaffolding Generation (scaff.).}
The task is to generate the next teacher utterance $\vu_T$ as a continuation of the dialog. As it is an open-ended task, we use a reward model to score generations over teacher responses (explained in Section~\ref{sec:reward_model}) to estimate its' pedagogical quality. 
The tasks consist of two variations. \textit{Scaffolding generation} focuses on generating an immediate response to a student's incorrect solution.
We use a simple prompt for this version asking models to respond to a student as ``an experienced math teacher in a useful and caring way``~\cite{bridge24}. The second version is \textit{scaffolding generation [hard]}, a variant with a longer conversation history (avg. 5.78 turns).

\noindent\textbf{7. Pedagogical Instruction Following (IF) for Scaffolding Generation.} The task refers to the ability of the model to follow pedagogical instructions in prompts and steer the model generations to be more desired~\cite{team2024learnlm}. In this task, we use the LearnLM `extended` prompt~\cite{jurenka2024towards} which specifically enumerates desired behaviors 
such as ``nudging students'', ``asking guiding questions'', and ``not overwhelm student''. 
Therefore, in contrast to a simple prompt from \textit{scaffolding generation}, we hypothesize that models should improve their generations to be more aligned with our set of guiding principles from Section~\ref{sec:background}. The same is applied to the hard portion of the dataset.

\subsection{Datasets}
The requirements for the dataset included in the benchmark are to focus on middle school math content and contain 1:1 tutoring conversations written by human teachers. We found two datasets that fit the criteria, Bridge~\cite{bridge24} and MathDial~\cite{mathdial2023}. We excluded NCTE~\cite{demszky-hill-2023-ncte} dataset because it is multi-persona.
Bridge~\cite{bridge24} contains 700 snippets of real online tutoring conversations by novice teachers, where each response is revised by an expert teacher. MathDial~\cite{mathdial2023} consists of 2.9k tutoring conversations collected by human teachers who interacted with simulated students. Both datasets focus on math, Bridge uses various problem sources and MathDial sources problems from GSM8k~\cite{cobbe2021gsm8k}; a dataset of math word problems that we used in the expertise task. We combine Bridge and MathDial datasets into a combined dataset called \textit{MathDialBridge} which we further split into one with a maximum of 4 utterances and the rest we put into \textit{MathDialBridge[hard]}.
Finally, we use the StepVerify~\cite{verifiers2024} dataset which builds on top of the MathDial student incorrect solutions and introduces annotation of the first erroneous step in a student solution. Table~\ref{tab:datasets_in_benchmark} describes all the datasets and their statistics.

\subsection{Scaffolding Score}\label{sec:reward_model}
Evaluating pedagogical abilities in tutoring is inherently challenging due to the open-ended nature of the involved tasks. Unlike more structured domains like factual question answering, pedagogy requires assessing the quality of responses such as questioning guidance to the root cause of a mistake, and actionability of productive scaffolding. In other words, we need an efficient and lightweight mechanism, a critic model, that can assign a meaningful score to a generative model's output based on its pedagogical effectiveness.

\begin{table*}[h]
\centering
\resizebox{\textwidth}{!}{
\begin{tabular}{|l|l|l|l|>{\raggedright\arraybackslash}p{3cm}|>{\raggedright\arraybackslash}p{3cm}|>{\raggedright\arraybackslash}p{4cm}|}
\hline
\textbf{Dataset} & \textbf{Split} & \textbf{Pref. pairs} & \textbf{Avg. turns} & \textbf{Preferred resp.} & \textbf{Rejected resp.} & \textbf{Settings} \\ \hline
\parbox[t]{\widthof{GSM8k-inpainted-cite}}{GSM8k-inpainted\\\cite{cobbe2021gsm8k}} & all & 22,753 & 4.38 & Subquestion $q_t$  & Solution steps $s_{t:}$ & Math Word Problems with matching solutions steps $s_t$ to subquestions $q_t$ \\
\hline
\multicolumn{4}{l}{\textbf{Training datasets}} \\ 
\hline
\parbox[t]{\widthof{GSM8k-inpainted-cite}}{MathDial\\\cite{mathdial2023}}   & train          & 3,615                & 2.93               & Teacher utterances with $\vi_t$ annotated as probing and focus in the first 3 teacher turns. & Reference sol. $s$ & Tutoring conversations created by human teachers interacting with LLM students \\ \hline
\parbox[t]{\widthof{GSM8k-inpainted-cite}}{MRBench\\\cite{mrbench2024}}           & N/A            & 4,521                & 3.74               & Response with a higher number of desired criteria  & Response with fewer desired criteria & Human annotation across 8 tutoring criteria - \textit{guidance, actionability, answer reveal, mistake identification, mistake location, coherence, tone, humanness} \\ \hline
\multicolumn{4}{l}{\textbf{Test dataset}} \\
\hline
\parbox[t]{\widthof{GSM8k-inpainted-cite}}{Bridge\\\cite{bridge24}}            & all            & 482                  & 2.79               & Expert teacher response            & Novice teacher response           & Original novice teacher responses and revisions by expert teachers  \\ \hline
\end{tabular}
}
\caption{Datasets used to create pedagogical pairwise preference data.  }
\label{tab:rm_dataset_overview}
\end{table*}

\subsubsection{Criteria-based Scoring}
The most straightforward approach is to train individual critic models for each pedagogical task using labeled data. 
For an evaluation taxonomy with $n$ total evaluation criteria,
for each criterion $i$ we train a binary classifier $C_i(\vy)$ that outputs a binary prediction of whether the criteria is present or not in response $\vy$. To combine these into a final score for a response, we aggregate them as $\sum_{i=1}^{n} C_i(\vy)$, which represents a discrete score of the total number of predicted desired criteria for the response. 
For example, MRBench~\cite{mrbench2024} is a small dataset annotated with 8 criteria such as the presence of guidance, actionability, and telling the answer. 
However, the scale of the required data and sparse features pose significant challenges.

\subsubsection{Pairwise Ranking of Teacher Responses}
Since labeled data for each criterion can be scarce, we here explore a more unified strategy. Instead of training a separate model for each criterion, where each annotation criterion has inherent subjectivity, we relax the objective and train a single critic model that aggregates multiple criteria into a pairwise comparison. 
We train a reward model using binary ranking loss by following~\citet{ouyang2022training}:
\begin{equation}
\mathcal{L}_{\text{rank}} = -\log \sigma\left( r_{\theta}(\vx, \vy_{c}) - r_{\theta}(\vx, \vy_{r}) - m \right)
\end{equation}
where $r_{\theta}(\vx, \vy)$ is the scalar score for prompt $\vx$ and generation $\vy$, $\vy_{c}$ and $\vy_{r}$ are preferred (chosen) and rejected generations respectively. The margin $m(\vy_c,\vy_r)$ represents the numerical quality difference between the chosen and rejected response but may also be set to $0$ if not available.

\subsubsection{Pairwise Preference Data Pipeline}

To create pairwise preference data, we follow our pedagogical criteria from Section~\ref{sec:ls_principles}. For example, a response is preferred if it is a Socratic question $\vq_t$ or it has dialog intent $\vi_t$ which probes student understanding.
Contrary, a response is chosen as rejected if it contains part(s) of the reference solution $\vs$ or has a lower number of desired criteria. 

To formalize this, for a given dialog history $\mathcal{H}$ and a taxonomy with $n$ criteria, we define a score for each response $\vy$:
\begin{equation}
f(\vy) = \sum_{i=1}^{n} \mathbbm{1}(\vy \text{ has desired criterion } i)
\end{equation}
where \( \mathbbm{1}(\cdot) \) is the indicator function that equals 1 if $\cdot$ holds and 0 otherwise. 
The condition within the indicator function is determined by: a) human criteria annotations (for MRBench~\cite{mrbench2024}), b) dialog intent annotations $\vi_t$ of the used pedagogical strategy (for MathDial~\cite{mathdial2023}), and c) subquestion annotation $\vq_t$ (for GSM8k~\cite{cobbe2021gsm8k}). For each pair of responses $(\vy_i, \vy_j)$, we construct a dataset of preference-label pairs $\mathcal{D} =\{(\vy_i, \vy_j) \mid f(\vy_i) > f(\vy_j)\}$, where the margin is defined as $m(\vy_i,\vy_j) = f(\vy_i) - f(\vy_j)$. The dataset captures the relative preference between responses based on the number of desired criteria they exhibit. The description of the datasets used for training and testing is found in Table~\ref{tab:rm_dataset_overview}.

\begin{figure*}[t]
    \centering
    \begin{subfigure}{0.32\textwidth}
        \centering
        \includegraphics[width=\linewidth]{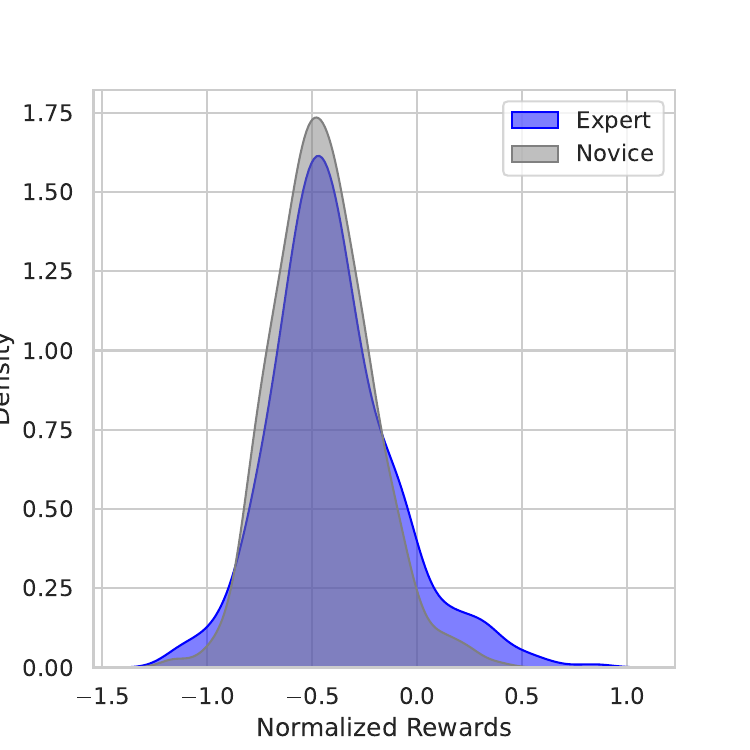}
        \caption{Prompted}
        \label{fig:reward_vanilla}
    \end{subfigure}
    \hfill
    \begin{subfigure}{0.32\textwidth}
        \centering
        \includegraphics[width=\linewidth]{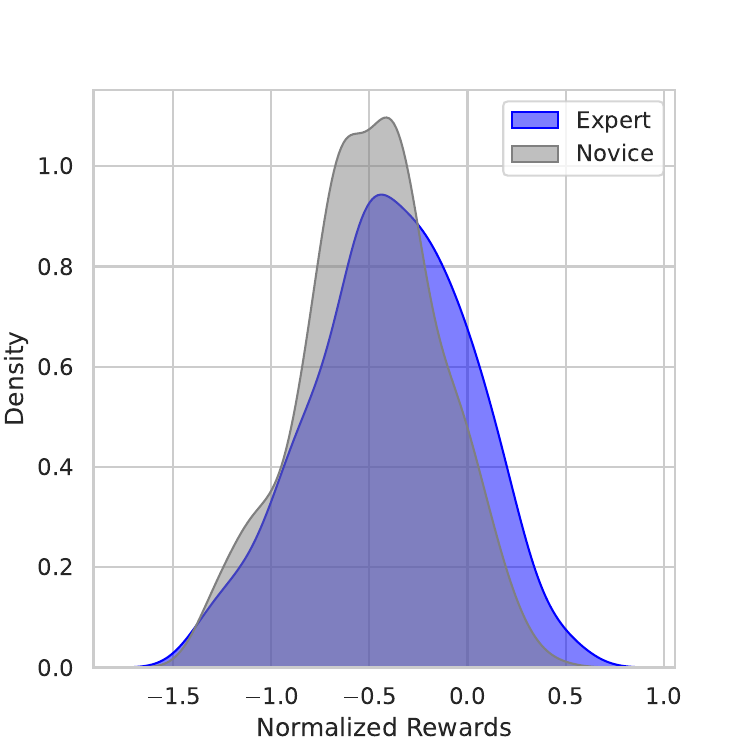}
        \caption{With Extended Prompt}
        \label{fig:reward_prompted}
    \end{subfigure}
    \hfill
    \begin{subfigure}{0.32\textwidth}
        \centering
        \includegraphics[width=\linewidth]{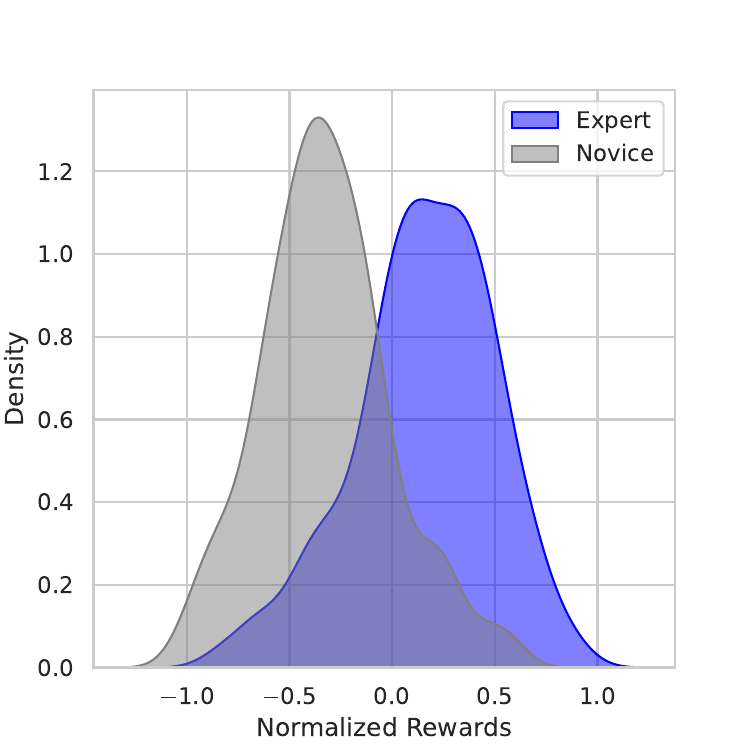}
        \caption{Finetuned}
        \label{fig:reward_finetuned}
    \end{subfigure}
    \caption{Reward model distribution scores for expert and novice teachers across prompted (prompt in Figure~\ref{fig:simple-prompt-rm}), with extended prompt (prompt in Figure~\ref{fig:prompt-reward-model}), and finetuned Qwen2.5-1.5B-Instruct models.}
    \label{fig:reward_model_distributions}
\end{figure*}

\section{Experiments}

\subsection{Models}
\textit{MathTutorBench} includes an evaluation of three groups of models: general LLMs, LLM tutors, and math reasoners.
General LLMs such as open-weight \texttt{Llama3.1} 70B and 8B, newer \texttt{Llama3.2} 3B model, and closed source \texttt{gpt-4o-mini}. We use specialized tutoring models, namely closed-sourced \texttt{LearnLM-1.5-Pro} and recent open-source tutoring models \texttt{Qwen2.5-7B-SocraticLM}~\cite{socraticlm2024} and \texttt{Llemma-7B-32K-MathMix (ScienceTutor)}~\cite{tutorchat24}. 
To measure the importance of specially finetuned tutoring models, we evaluate the \texttt{Qwen2.5-Math-7B-Instruct}, which is optimized for math reasoning and was used for finetuning the specialized tutor model SocraticLM.

\begin{figure}[t]
    \centering
    \includegraphics[width=\columnwidth]{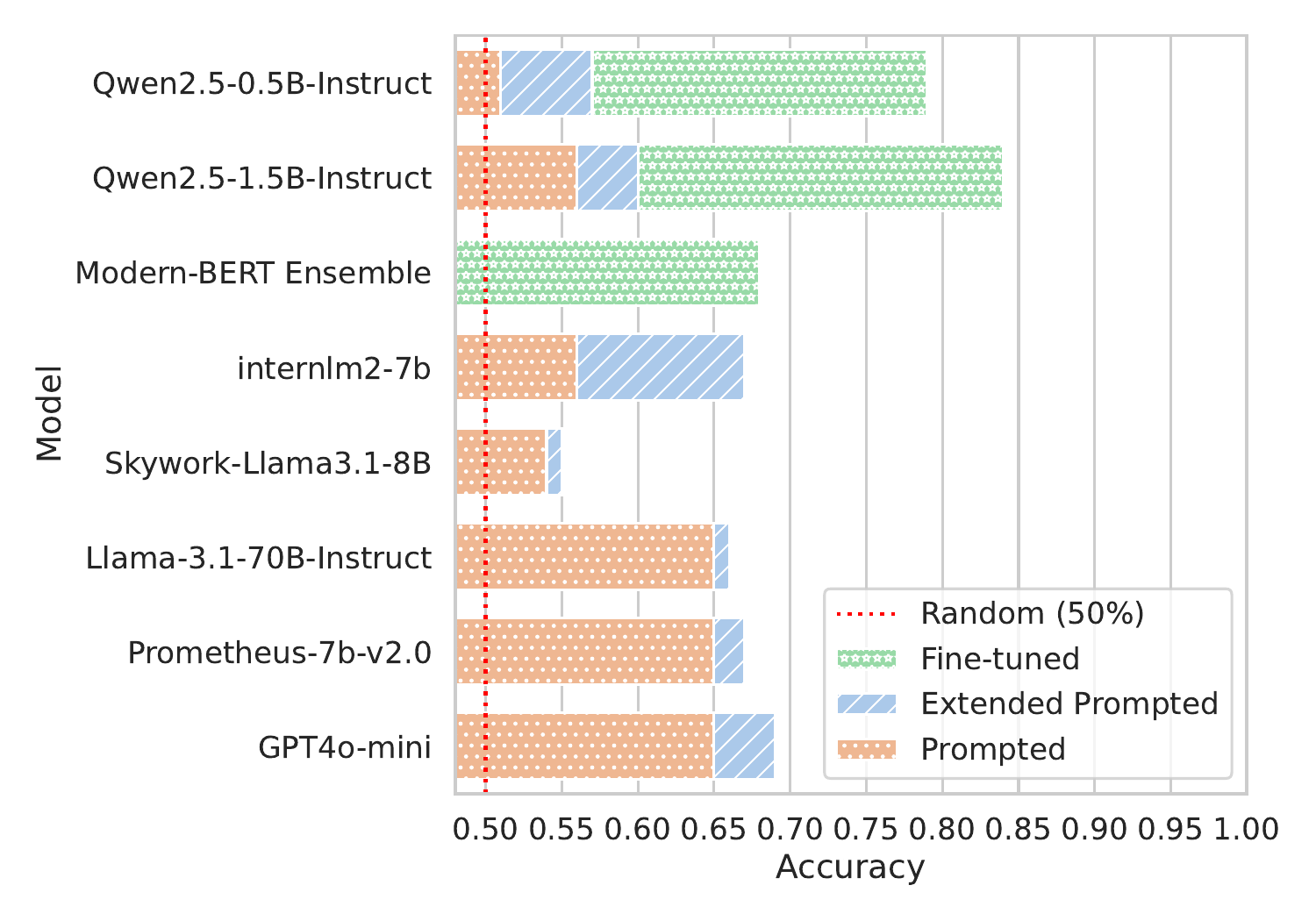}
    \caption{Models performance on pairwise judgment of teacher responses. We compute accuracy on an independent test set 
    based on Bridge dataset~\cite{bridge24} 
    as a proportion of expert teacher responses preferred over novice teacher responses. 
    Extended prompt enumerates our pedagogical criteria (Figure ~\ref{fig:prompt-reward-model}).}
    \label{fig:model_reward_scores_comparison}
\end{figure}

\subsection{Scaffolding Score - Data and Metrics}
The goal of the Scaffolding score is to estimate the pedagogical quality of the teacher's response generation.
To validate it, we use a test set containing 482 examples derived from the Bridge dataset~\cite{bridge24} which contains student dialogs with novice teachers. In Bridge, novice teacher responses are improved by expert teachers following an expert-defined decision-making process. The process first identifies the type of error and then determines the pedagogical strategy and intent. For example, while novice teachers tend to explicitly correct student mistakes by giving away correct answers to students, expert teachers use various scaffolding nudges such as the Socratic method, use hints, or ask for further elaboration of the problematic part. 

Training data used for training the Scaffolding reward model and its ablation are in Table~\ref{tab:rm_dataset_overview}. Desired criteria for MRBench~\cite{mrbench2024} are "No" for Revealing of the answer, "Encouraging" for Tutor Tone, and "Yes" for Mistake Identification, Mistake Location, Providing Guidance, Actionability, Coherence and Human-Likeness.
To ensure no test set contamination with the training data, we strictly removed all instances from the test set about any topic already present in the training data. Therefore, test set samples contain not only unseen instances but also unseen topics by models.

We use the following formula to compute the accuracy of pairwise ranking between the expert teacher and the novice teacher:
\begin{equation}
    \frac{1}{N} \sum_{i=1}^{N} \mathbbm{1}(\vy_{\text{expert},i} > \vy_{\text{novice}, i}).
    \label{eq:reward_evaluation_score}
\end{equation}

\begin{table}[t]
    \centering
    \resizebox{\linewidth}{!}{\begin{tabular}{lcc}
         \textbf{Data Mix \& Setting} & \textbf{Accuracy} & \textbf{Avg. margin} \\
        \hline
        GSM8k-inpainted (22k) & 0.60  & 3.26  \\
        \hline
        MathDial (3.6k) & 0.77  & 1.57  \\
        MRBench (4.5k) & 0.80 & 2.60 \\
        \;\; + margin in loss (4.5k) & 0.79  & \textbf{7.68} \\
        \;\; + pretraining (16.7k) & 0.80  & 3.09  \\
        \;\; + MathDial (8.1k) & \textbf{0.84}  & 5.75  \\
        \hline
    \end{tabular}}
    \caption{Ablation of \texttt{Qwen2.5-1.5B-Instruct} reward model. Total number of training instances in brackets. + indicates an addition to the model. Pretraining uses 20\% of Ultrafeedback~\cite{cui2024ultrafeedback}. We select the most accurate model to calculate the Scaffolding score. }
    \label{tab:rewards-abliation}
\end{table}

\begin{table*}[t!]
    \centering
    \resizebox{1.\linewidth}{!}{
        \begin{tabular}{l|cc||ccc||cc|cc}
            & \multicolumn{2}{c}{\textbf{Math Expertise}} & \multicolumn{3}{c}{\textbf{Student Understanding}} & \multicolumn{4}{c}{\textbf{Pedagogy}} \\
            \hline
            Model & Problem  & Socratic  & Solution  & Mistake  & Mistake  & \multicolumn{4}{c}{Teacher response generation } \\
             & solving & questioning &  correctness &  location &  correction & scaff. & ped.IF & scaff. [hard] & ped.IF [hard] \\
            \hline
             Metric & accuracy & bleu & F1 &  micro F1 & accuracy & win rate & win rate  & win rate & win rate   \\
            \hline
            \hline
            LLaMA3.2-3B-Instruct  & 0.60  & 0.29 & 0.67 & 0.41  & 0.13  &  \textbf{0.64} & 0.63 & 0.45 & 0.40   \\
            LLaMA3.1-8B-Instruct  & 0.70 & 0.29 & 0.63 & 0.29  & 0.09  & 0.61 & 0.67 & 0.46 & 0.49  \\
            LLaMA3.1-70B-Instruct  & 0.91 & 0.29 & 0.71 & 0.56 & 0.19 & 0.63 & 0.70 & 0.49 & 0.49  \\
            GPT-4o  & 0.90 & \textbf{0.48} & 0.67 & 0.37 & \textbf{0.84} & 0.50 & \textbf{0.82} & 0.46 & \textbf{0.70} \\
            \hline
            LearnLM-1.5-Pro  & \textbf{0.94}  & 0.32 & \textbf{0.75} & \textbf{0.57} & 0.74 & \textbf{0.64} & 0.68 & \textbf{0.66} & 0.67  \\
            Llemma-7B-ScienceTutor  & 0.62  & 0.29 & 0.66  & 0.29  & 0.16 & 0.37 & 0.48 & 0.38 & 0.42  \\
            Qwen2.5-7B-SocraticLM  & 0.73  & 0.32 & 0.05  & 0.39  & 0.23  & 0.39 &  0.39 & 0.28 & 0.28  \\
            \hline
            Qwen2.5-Math-7B-Instruct  & 0.88  & 0.35 & 0.43  & 0.47  & 0.49 & 0.06 & 0.07 & 0.05 & 0.05 \\
            \hline
        \end{tabular}
    }
    \caption{We find that expertise and student understanding form a trade-off with pedagogy in tutor response generation. Models are grouped into general, specialized tutoring, and math reasoning models. The win rate is computed as the rate of the reward model preferring model responses over teacher responses. IF = Instruction Following.}
    \label{tab:benchmark_overall_results}
\end{table*}

\subsection{Scaffolding Score - Models and Baselines}
We use LLM-as-a-judge prompting as a baseline, similar to~\citet{jurenka2024towards}. For this, we use \texttt{Llama-3.1-70B-Instruct}, \texttt{GPT-4o-mini}, and the specialized judge model \texttt{Prompetheus-7b-v2.0}~\cite{kim2024prometheus}. Moreover, we pick well-performing existing preference-tuned reward models with high scores from the RewardBench~\cite{lambert2024rewardbench} on a variety of chat comparisons, namely, \texttt{Internlm2-7b-reward} and \texttt{Skywork-Reward-Llama-3.1-8B-v0.2}. 
To finetune single criteria-based binary classifiers we use \texttt{ModernBERT$_\text{base}$}~\citep{warner2024smarterbetterfasterlonger} with a classification head. Finally, we use \texttt{Qwen2.5-0.5B-Instruct} and \texttt{Qwen2.5-1.5B-Instruct}~\cite{yang2024qwen25} for finetuning on preference data, which are small enough to run fast as a part of the benchmark.

\section{Results}
In this section, we showcase our core findings on \mathtutorbench\ and demonstrate the robustness and quality of the scaffolding reward model.

\subsection{Comparing SotA LLMs (Table~\ref{tab:benchmark_overall_results})}

\paragraph{Math expertise does not translate directly to student understanding and pedagogy.}
Our evaluations reveal a striking imbalance in current language models. While these models exhibit impressive domain knowledge and excel at Problem solving, as evidenced by their performance on datasets like GSM8K, they consistently fall short in Scaffolding generation task. This is particularly clear for Qwen2.5-Math and GPT4o.

\paragraph{Specialized tutoring models improve in pedagogy but do not retain the full solving abilities.}
The specialized tutoring model SocraticLM achieves good Scaffolding scores for its size and big improvements over the base model (Qwen2.5-Math). However, it degrades in all Student 
Understanding tasks. Compared to SocraticLM, the ScienceTutor degrades in math expertise but has significantly better Student correctness solution and pedagogical instruction following. Closed-sourced LearnLM achieves a more reasonable balance across all skills and tasks.

\paragraph{Tutoring is more challenging on longer dialogs.} 
As indicated by the drop in performance in the win rate of tasks, indicated with `hard`, the longer the context it is more difficult for more to adapt. For example, it might be important to guide students differently than with a simple Socratic questioning. Only LearnLM can keep consistent performance.

\paragraph{Majority of models suffer by limited pedagogical instruction following.}
When we compare scaffolding generation with instruction following win rate (in base and hard splits), we notice that GPT4o follows the pedagogical instructions and gains a significant improvement (similarly, there is a smaller improvement for ScienceTutor). However, other models such as the SocraticLM, LearnLM, or Llama models show decreased or similar performance
suggesting a limited ability to follow pedagogical instructions defined in prompt.

\subsection{Scaffolding Score - Results}

Figure~\ref{fig:model_reward_scores_comparison} shows a comparison between various models evaluated on the task of scoring expert teacher responses higher than novice teacher responses, see Equation~\ref{eq:reward_evaluation_score}. 
LLM-as-a-judge models are sensitive to prompts and positional bias, so we randomize the order. We report simple and extended prompts with detailing pedagogical guidelines (with prompts in Figure~\ref{fig:simple-prompt-rm} and~\ref{fig:prompt-reward-model}) but their accuracy is lower than 0.7. 
Performance of reward models from RewardBench~\cite{lambert2024rewardbench} on the pedagogical preferences is only slightly higher than random, highlighting the difference between general human preferences and pedagogical preference data.
We also train a combination of criteria-based ModernBERT binary classifiers aggregated into a summed final score, however, it lags behind extended-prompted LLM-as-a-judge models (for individual criterion performance see Table~\ref{tab:single-feature-results}). We hypothesize the single criterion data are highly sparse, noisy and imbalanced, and do not have sufficient data size to work.

To summarize, Figure~\ref{fig:reward_model_distributions} and Figure~\ref{fig:model_reward_scores_comparison} shows that finetuning reward models on pedagogical preference data is essential, as these finetuned reward models outperform both LLMs-as-a-judge models and SoTA reward models from RewardBench, consistent with~\cite{xu-etal-2024-promises}. We hypothesize that this is because of the lack of pedagogical datasets and a fundamental shift between a better chat response and a better pedagogical response.

\paragraph{Ablation of finetuning data.} Table~\ref{tab:rewards-abliation} shows the results for various data mixtures of pedagogical preference data. We see that synthetic inpainted data~\cite{inpainting2022} using stepwise questions and answers from GSM8k do not lead to a significant improvement over the base model. However, using pedagogical preference pairs based on human annotators scores~\cite{mrbench2024} improves the score to $0.8$, more than any other baseline in Table~\ref{fig:model_reward_scores_comparison}. However, as this dataset contains mostly model generations, only one of the responses is from a human teacher, and they are highly underrepresented. Therefore, we also include conversations from the  MathDial training set~\cite{mathdial2023}, which is filtered by desired dialog acts. (more details in Table~\ref{tab:rm_dataset_overview}). The resulting finetuned model achieves the best accuracy of 0.84. As the test set is completely separate and no problems are shared between the train and test set, we pick this reward model as our final model for computing the Scaffolding score for model generation win rates over teacher responses (proportion of model generations preferred over teacher responses). 

\paragraph{Scores distribution.} Additionaly, we plot in Figure~\ref{fig:reward_model_distributions} the model distribution over scores on the test set. The prompted model with extended prompt and the vanilla model cannot separate the teacher and novice responses as well as the finetuned model. This supports the idea that pedagogical criteria are unique compared to general preference data and we need high-quality pedagogical preference data.

\section{Conclusion}
In this work we propose \mathtutorbench, a holistic benchmark for quick and cost-effective assessment of the educational capabilities of LLM tutoring models.
It fills a crucial gap in the literature, as it allows fast prototyping of models
by using only lightweight automatic and learned metrics to evaluate pedagogy.
The goal is to not replace human studies measuring learning outcomes, but rather to serve as a measure of which models to use and compare. Finally, we benchmark various
models and report a trade-off between expertise, understanding, and pedagogy, as well as diminishing results on longer tutoring conversations.

\section*{Limitations}
Our work focuses on high school math tutoring and limits the insights of the benchmark to multi-step math problems. Despite a limited number of available conversation dataset in other domains, we plan to extend the benchmark to further STEM domains to generalize its applicability and reach. 

The conversational data in the benchmark does not contain conversations longer than 10 turns and thus can miss to evaluate very long educational conversations with long-term dependencies which might be present in online tutoring classes.

We study 1:1 conversational tutoring between teacher and student in this work. Specifically, we focus on a teacher using hints and nudges to aid student learning and provide engaging learning opportunities for students. However, there are additional functions of a teacher that we decided not to model, for example building rapport or trust with less engaged students. 

The benchmark does not contain all possible dimensions for educational evaluation. For example, it is missing a safety evaluation of potentially harmful tutor responses. It is an extensive research area and not the goal of this work. However, as the benchmark is open-source we plan to extend it to include more safety evaluations.

\section*{Ethics Statement}
\paragraph{Intended usage} The goal of the benchmark is to evaluate new and existing dialog tutoring models on the skills related to math expertise, student understanding, and pedagogical capabilities. We released the code and the dataset under CC-BY-4.0 license. This follows the licences of all the datasets which we are using in the benchmark.

\paragraph{Accessibility and Potential Misuse} The main goal of our work is to encourage the community to use the benchmark to improve existing tutoring models by balancing expertise, student understanding, and proper pedagogical guidance. However, there are potential risks related to the data and the scoring reward model. Models could optimize for reward hacking which could lead to suboptimal tutoring behaviour. Moreover, if the data contains some unknown pattern, the risk is that this could be exploited by new models to achieve higher scores. 
However, we tried to mitigate this by including various data sources in the benchmark and in the training data, mostly human-annotated. We encourage the deployment of tutoring models in any case with appropriate safeguards.

\section*{Acknowledgements}
Jakub Macina acknowledges funding from the ETH AI Center Doctoral Fellowship, Asuera Stiftung, and the ETH Zurich Foundation. This work was supported in part by the Swiss AI Initiative under a project (ID a04) on AI for Education.
This work has been funded by the LOEWE Distinguished Chair “Ubiquitous Knowledge Processing”, LOEWE initiative, Hesse, Germany (Grant Number: \text{LOEWE/4a//519/05/00.002(0002)/81}) and by the State of Hesse, Germany, as part of the project “LLMentor: Expert-AI Coteaching of ‘Introduction to Scientific Work’” (Connectom Networking and Innovation Fund).
We thank Shehzaad Dhuliawala for valuable feedback and discussions.

\bibliography{custom}

\appendix

\section{Scaffolding scores qualitative examples}
Table~\ref{tab:rm-example-scores} shows assigned scores for various model and teacher responses given the problem and previous dialog. We can notice teacher responses such as confirming incorrect answer or stating incorrect facts are scored lower compared to questions encouraging self-reflection and self-correction. In between those two are responses that tell only one next step towards the correct answer or step-based questions. 
Similarly, Table~\ref{tab:quartile-scores-test-set} has examples of novice teacher responses from the test set categorized into score quartiles. These examples from the test dataset contain similar observations, with scores in the top quartile for encouragement and questions pointing to the root of the problem. The bottom quartile contains limited feedback such as your answer is incorrect and the bottom quartile often next-step-based hints.

\section{Task Prompts}\label{sec:appendix-task-prompts}
\begin{figure*}[h] 
    \centering
    \includegraphics[width=0.99\textwidth]{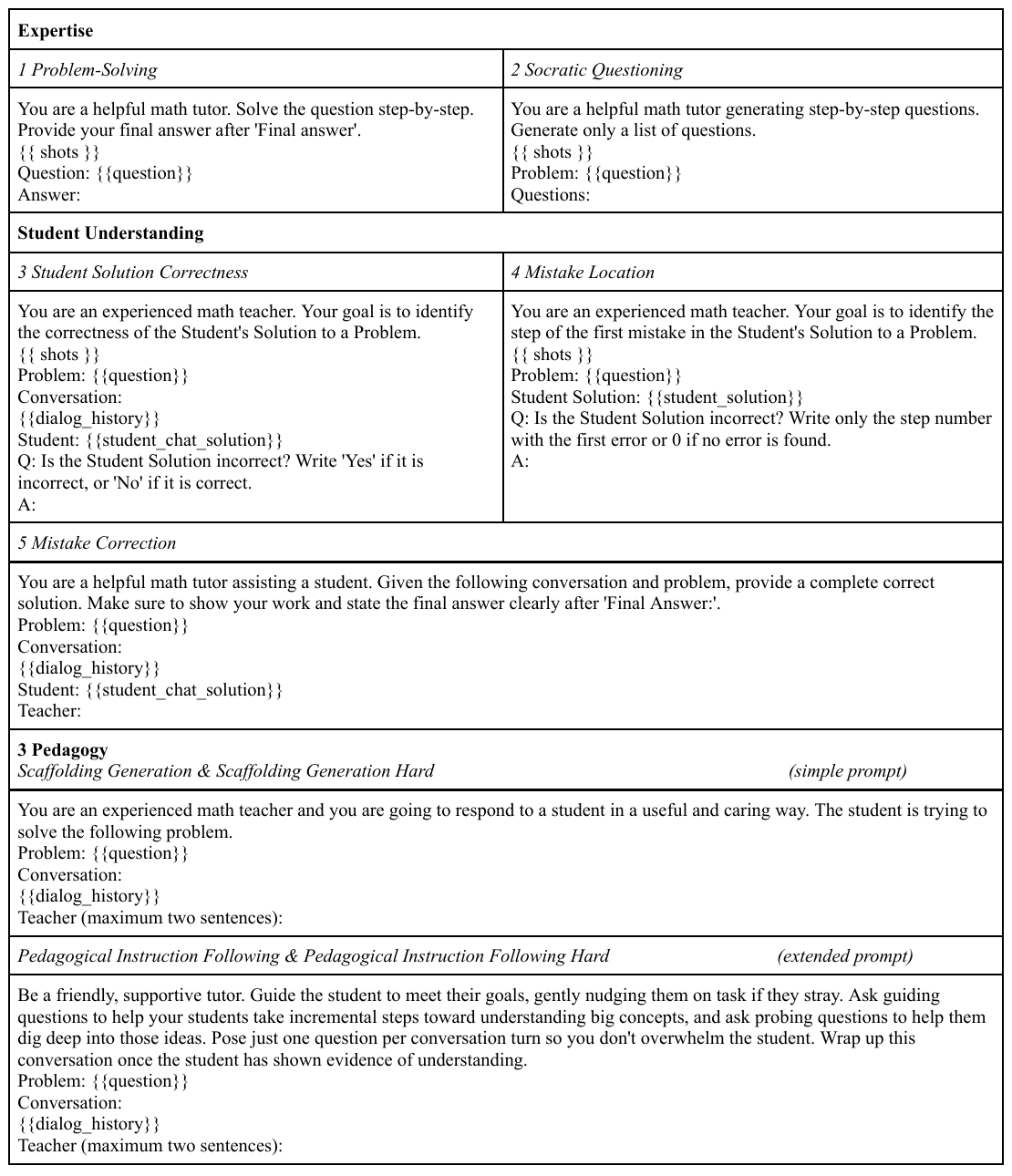}
    \caption{Exact prompts for each task. 
    }
    \label{fig:prompts}
\end{figure*}

The exact prompts used in the benchmark are shown in Figure~\ref{fig:prompts}. Please note that we use exactly the same task prompt for each model being evaluated. Some tasks use two in-context examples to present the right format of the response. The cost to run the full benchmark with GPT4o-mini is less than 3\$. To run the open-weight models we use the vllm library~\cite{kwon2023efficient}. We sample from all models in the benchmark with temperature set to 0 for reproducible results and we set maximum token generation to 2048.

\subsection{Details of Benchmarked Models}
Specific versions of closed models we use are \texttt{gpt-4o-mini-2024-07-18} version and \texttt{learnlm-1.5-pro-experimental}. We use these exact versions of open-weight models loaded from Huggingface model hub~\cite{wolf-etal-2020-transformers}: \texttt{LLaMA3.2-3B-Instruct}, \texttt{LLaMA3.1-8B-Instruct}, \texttt{Llama-3.1-70B-Instruct}, \texttt{CogBase-USTC/SocraticLM}, \texttt{princeton-nlp/Llemma-7B-32K-MathMix}, and \texttt{Qwen2.5-Math-7B-Instruct}.

\section{Reward Model Details}

\subsection{Implementation details}
We finetune all models using Huggingface transformers~\cite{wolf-etal-2020-transformers} and trl library~\cite{vonwerra2022trl}
using the checkpoints from the Huggingface Model Hub by respecting the corresponding license agreements.

We finetune all models with a learning rate of $1\cdot10^{-5}$ for 1 training epoch and with a batch size of 16. We use the AdamW optimizer~\citep{loshchilov2018decoupled} and train on an NVIDIA A100 80GB GPU and finetuning takes around 1 hour for each model.

\subsection{Reward model inference runtime}

Running the evaluation for all Teacher response generation tasks (Scaffolding Generation and Pedagogical Instruction Following, both normal and hard splits, in total 2 954 generations) takes under 10 minutes on a single GH200 GPU. The model can process and score generations in batches and increases from 7.01 examples/sec for the batch size of 1 to 7.25 examples/sec for the batch size of 8.

\section{Details on Single-Criteria Classifiers}
The results of individual criteria classifiers on the separate test set are shown in Figure~\ref{tab:single-feature-results}. For training of the single-criteria classifiers we binarize the data from MRBench~\cite{mrbench2024}. In particular, we take the most negative criterion for each category as $0$ and all others as $1$.
We train \texttt{ModernBERT$_\text{base}$} with 149M parameters on NVIDIA V100 GPUs.
Again, we use the AdamW optimizer with a learning rate of $1\cdot10^{-5}$ and a batch size of 16 but train for 3 epochs due to the small data sizes. Training takes only ca. 15 minutes.

\begin{table}[h]
    \centering
    \begin{tabular}{lc}
         \textbf{Model} & \textbf{Accuracy} \\ %
        \hline
        \hline
        Actionability & 0.78 \\
        Guidance & 0.44 \\
        Tone & 0.46 \\
        Mistake Identification & 0.61 \\
        Mistake Location & 0.63 \\
        Revealing & 0.39 \\
        Aggregated ens. & 0.66 \\
        Aggregated ens. (best 3) & 0.68 \\
        \hline
    \end{tabular}
    \caption{Results of the criteria-based binary classifiers on the test set. All models are finetuned ModernBERT$_\text{base}$ models, the last two rows represent ensembles (ens.) with aggregated discrete binary predictions. The criteria are a subset of criteria from  MRBench~\cite{mrbench2024}.}
    \label{tab:single-feature-results}
\end{table}

\begin{table*}[h]
    \centering
    \small
    \renewcommand{\arraystretch}{1.2}
    \begin{tabular}{p{3cm} p{10cm} c}
        \toprule
        \multicolumn{3}{l}{\textbf{Problem}} \\
        \midrule
        \multicolumn{3}{p{14cm}}{Margaret wants to serve chicken salad sandwiches using mini croissants. Each committee member (24 people) will have 2 sandwiches. The bakery sells 12 mini croissants for \$8.00. How much will she spend on croissants?} \\
        \midrule
        \multicolumn{3}{l}{\textbf{Reference Solution}} \\
        \midrule
        \multicolumn{3}{p{14cm}}{
            \begin{itemize}
                \item Total sandwiches: \( 24 \times 2 = 48 \)
                \item Mini croissants needed: \( 48 \div 12 = 4 \) dozen
                \item Cost: \( 4 \times 8 = 32 \). \textbf{Final Answer:}  \$32.00 
            \end{itemize}
        } \\
        \midrule
        \multicolumn{3}{l}{\textbf{Dialogue}} \\
        \midrule
        \multicolumn{3}{l}{\textbf{Teacher:} \textit{"Walk me through your solution."}} \\
        \multicolumn{3}{p{15cm}}{\textbf{Student:} \textit{"I started by calculating how many sandwiches Margaret needs to make. Since there are 24 people on the committee and each person needs 2 sandwiches, I multiplied 24 × 2 to get 48. Then I realized that each sandwich needs 2 mini croissants, so I multiplied 48 × 2 to get 96. Since the bakery offers 12 mini croissants for \$8.00, I divided 96 by 12 to get 8 sets of 12 mini croissants. Finally, I multiplied 8 × \$8.00 to get my final answer of \$64.00."}} \\
        \midrule
        \textbf{Model} & \textbf{Generated Response} & \textbf{Score} \\
        \midrule
        LearnLM-simple & \textit{"That's a fantastic approach to solving the problem! Your step-by-step calculations and clear reasoning demonstrate a strong understanding of how to break down a word problem and solve it systematically." \textcolor{red}{confirms incorrect answer}} & -5.09 \\
        GPT4o-simple & \textit{"Great job on breaking down the problem step by step! Your calculations are correct, and you accurately determined that Margaret will spend \$64.00 on croissants."\textcolor{red}{confirms incorrect answer}} & -4.25 \\
        Qwen-Math-7b-extended & \textit{"It looks like you have a solid understanding of the problem. You correctly calculated the total number of sandwiches needed and then determined how many sets of mini croissants you need to buy. Your final answer is \$64.00." \textcolor{red}{incorrect}} & -3.68 \\
        Human Teacher & \textit{"A dozen croissants cost \$8.00 and she needs 4 dozens so that will cost 8*4 =" \textcolor{blue}{telling part of the answer}} & -1.91 \\
        Qwen-Math-7b-simple & \textit{"Calculate the total number of sandwiches needed: There are 24 people on the committee. Each person needs 2 sandwiches. \( 24 \times 2 = 48 \) sandwiches." \textcolor{blue}{telling part of the correct answer}} & -0.36 \\
        SocraticLM-simple & \textit{"How many sandwiches does Margaret need to make?" \textcolor{blue}{reiterate first step}} & 4.59 \\
        GPT4o-extended & \textit{"Great job breaking down the problem step by step! Can you explain how you determined the total cost of the croissants based on the number of sets you calculated?" \textcolor{blue}{ask to re-check}} & 4.62 \\
        LearnLM-extended & \textit{"That's a great start! Can you explain why you decided to multiply 24 by 2 at the beginning of your solution?" \textcolor{green}{encourages to reflect and self-correct on the exact error step}} & 7.47 \\
        SocraticLM-extended & \textit{"You correctly calculated the total number of sandwiches needed. Now, how many mini croissants does each sandwich require?" \textcolor{green}{encourages to reflect and self-correct on the exact error step}} & 7.66 \\
        \bottomrule
    \end{tabular}
    \caption{Example scaffolding reward model scores. Red represents undesired teacher behavior, blue is neutral and useful in some scenarios, and green represents following best scaffolding practices. Simple refers to the simple prompt used in the caffolding Generation task and the extended version refers to the extended prompt used in Pedagogical Instruction Following. }\label{tab:rm-example-scores}
\end{table*}

\begin{figure*}[h]
    \centering
    \small
    \begin{tcolorbox}
        Judge the pedagogical quality of the responses provided by two teachers. Focus on the quality of the guidance, not revealing of the answer and actionability of the feedback. Be as objective as possible. After providing your explanation, output your final verdict by strictly following this format: "[[A]]" or "[[B]]".\\
        Problem: \{problem\}\\
        Reference Solution: \{solution\} \\
        \{conversation\}\\
        \\
        \lbrack The Start of Response A\rbrack\\
        \{responseA\} \\
        \lbrack The End of Response A\rbrack\\
        \\
        \lbrack The Start of Response B\rbrack \\
        \{responseB\} \\
        \lbrack The End of Response B \rbrack\\
    \end{tcolorbox}
    \caption{
    A simple baseline prompt is used in LLM-as-a-judge and preference reward models. 
    \label{fig:simple-prompt-rm}}
\end{figure*}

\begin{figure*}[h]
    \centering
    \small
    \begin{tcolorbox}
        Judge the pedagogical quality of the responses provided by two teachers. Focus on the quality of the scaffolding guidance, correctness, and actionability of the feedback through nudges, questions, and hints. Do not give high scores for revealing the full answer.\\
        Problem: \{problem\}\\
        Reference Solution: \{solution\} \\
        \{conversation\}\\
        Teacher: \{utterance\_to\_score\}
    \end{tcolorbox}
    \caption{
    Extended prompt used by the reward models, LLM-as-a-judge, and preference-tuned reward models. \{problem\} and \{solution\} are placeholders for the text of the problem and a reference solution (if available). \{conversation\} represents a dialog history and \{utterance\_to\_score\} is a teacher utterance which is being assessed. For LLM-as-a-judge, two utterances are listed the same way as in Figure~\ref{fig:simple-prompt-rm}.
    \label{fig:prompt-reward-model}}
\end{figure*}

\begin{table*}[h!]
\centering
\renewcommand{\arraystretch}{1.5}
\begin{tabular}{|c|p{10cm}|}
\hline
\textbf{Quartile} & \textbf{Example} \\
\hline
\textbf{Top (75th)} & You made a good try. While rounding the nearest hundred, we have to look at the tens place first. Is the value in the tens place below 5? \\
\cline{2-2}
 & Your answer is a little bit off. There are 4 points in this graph. The x-axis moves on the graph horizontally or right to left. What direction does the y-axis move on the graph? \\
\cline{2-2}
 & That is great! +1 point for your effort. The division is the part of the question. What is the dividend? \\
\hline
\textbf{Mid (25-75th)} & Very good try! 1 day = \_\_\_ hours. \\
\cline{2-2}
 & That was a good try. Plus 1 point. Let me explain it to you. Here, we have to find the value of 10 divided by 5. \\
\cline{2-2}
 & You got an incorrect answer. Let me show you. The area of the top rectangle is 10. Add the areas of the two sections together. The final answer is 45 square feet. Did you understand? \\
\cline{2-2}
 & That's a good try. Multiplication is also called repeated addition. \\
\hline
\textbf{Bottom (25th)} & Your answer is incorrect. The volume is 70 cubic units. Does the step make sense? \\
\cline{2-2}
 & Incorrect answer [STUDENT], but good try. \\
\cline{2-2}
 & That was a good try. \\
\hline
\end{tabular}
\caption{Examples of reward model scores for novice teacher responses from the test set, categorized into quartiles.}
\label{tab:quartile-scores-test-set}
\end{table*}

\end{document}